\documentclass[10pt,twocolumn,letterpaper]{article}

\usepackage{wacv}              % To produce the CAMERA-READY version
\usepackage{graphicx}
\usepackage{amsmath}
\usepackage{amssymb}
\usepackage{booktabs}
\usepackage{algorithm}
\usepackage{algorithmic}
\usepackage{multirow} 

\usepackage[pagebackref,breaklinks,colorlinks]{hyperref}

\usepackage[capitalize]{cleveref}
\Crefname{table}{Table}{Tables}
\crefname{table}{Tab.}{Tabs.}

\def\wacvPaperID{31} % *** Enter the WACV Paper ID here
\def\confName{WACV}
\def\confYear{2025}

\begin{document}

\title{Confident Pseudo-labeled Diffusion Augmentation  for Canine \\Cardiomegaly Detection}

\author{Shiman Zhang, Lakshmikar Reddy Polamreddy, Youshan Zhang\\
Graduate Computer Science and Engineering Department\\
Katz School of Science and Health, Yeshiva University\\
205 Lexington Avenue, New York, NY 10016\\
{\tt\small szhang6@mail.yu.edu, lpolamre@mail.yu.edu, youshan.zhang@yu.edu}
%\and
%Lakshmikar\\
%Katz School of Science and Health, Yeshiva University\\
%\and
%Youshan Zhang\\
%Katz School of Science and Health, Yeshiva University\\
}
\maketitle

%%%%%%%%% ABSTRACT
\begin{abstract}
Canine cardiomegaly, marked by an enlarged heart, poses serious health risks if undetected, requiring accurate diagnostic methods. Current detection models often rely on small, poorly annotated datasets and struggle to generalize across diverse imaging conditions, limiting their real-world applicability. To address these issues, we propose a Confident Pseudo-labeled Diffusion Augmentation (CDA) model for identifying canine cardiomegaly. Our approach addresses the challenge of limited high-quality training data by employing diffusion models to generate synthetic X-ray images and annotate Vertebral Heart Score key points, thereby expanding the dataset. We also employ a pseudo-labeling strategy with Monte Carlo Dropout to select high-confidence labels, refine the synthetic dataset, and improve accuracy. Iteratively incorporating these labels enhances the model’s performance, overcoming the limitations of existing approaches. Experimental results show that the CDA model outperforms traditional methods, achieving state-of-the-art accuracy in canine cardiomegaly detection. The code implementation is available at \url{https://github.com/Shira7z/CDA}.

\end{abstract}

% This project leverages artificial intelligence (AI) to predict Vertebral Heart Score (VHS) from canine chest X-rays, enabling reliable detection of cardiomegaly. 
% Synthetic data, generated through advanced generative models and manually annotated by the researcher, significantly improve the robustness of the AI system.
% Expected outcomes include a highly accurate AI model capable of automated VHS prediction, filling a critical gap in veterinary diagnostics. This innovation not only enhances early detection of canine cardiomegaly but also demonstrates the potential of generative models and pseudo-labeling in advancing medical imaging applications for veterinary care.
%%%%%%%%% BODY TEXT
\section{Introduction}
\label{sec:intro}

Cardiomegaly, or the abnormal enlargement of the heart, is a critical factor in cardiac-related morbidity and mortality, affecting both human and animal health. In veterinary medicine, cardiomegaly is commonly associated with degenerative heart diseases in dogs, making early detection of this condition essential for improving health outcomes and quality of life. Recent advances in artificial intelligence (AI) and deep learning (DL), particularly with convolutional neural networks (CNNs), have shown considerable promise in the medical imaging field~\cite{Oh2023, Lee2013, Sohn2020}. These technologies offer the potential for autonomous and efficient detection of complex patterns in medical images, assisting radiologists in accurately diagnosing conditions across various imaging modalities. Nevertheless, the adoption of DL in veterinary diagnostics, especially for conditions like cardiomegaly, has yet to reach its potential compared to human medicine, where AI-based diagnostic aids have seen wider acceptance.

Applying AI technologies to cardiomegaly assessment for dogs can streamline the diagnostic process, significantly reducing time and costs. By automating aspects of diagnosis, AI systems can ease the workload on veterinary professionals, who often rely on traditional, labor-intensive methods. These methods, though familiar, involve subjective measurements that vary between clinicians, such as the vertebral heart score (VHS), a standard diagnostic metric in veterinary cardiology. However, manual VHS calculation is time-consuming and prone to human error, with variable results due to differences in identifying key anatomical points. The challenge of standardizing and optimizing cardiomegaly assessment highlights the need for AI models that can provide both accuracy and interpretability, bridging the gap between DL technology and practical clinical applications.

To address the challenges of limited training data and enhance diagnostic precision, pseudo-labeling techniques have been integrated into AI-based pipelines for VHS prediction. Pseudo-labeling involves leveraging high-confidence predictions from a pre-trained model to annotate unlabeled data, which is then incorporated into subsequent training cycles to iteratively improve model performance~\cite{Lee2013}. This approach is particularly useful in veterinary applications, where labeled datasets are often scarce. For example, Sohn et al.~\cite{Sohn2020} demonstrated that pseudo-labeling, combined with uncertainty-based selection, significantly boosts model robustness in medical imaging tasks. We apply this technique to canine cardiomegaly assessment by using predictions from the best-performing EfficientNetB7~\cite{tan2019efficientnet} model to pseudo-label an unlabeled dataset. High-confidence outputs, selected using Monte Carlo Dropout~\cite{gal2016dropout} for uncertainty estimation, are then incorporated into the training set, thereby enhancing the model’s generalizability.

Additionally, synthetic data generation has emerged as a powerful tool to address the scarcity of labeled datasets in veterinary diagnostics. Diffusion models, particularly advancements such as Latent Diffusion Models (LDMs)~\cite{rombach2022high}, have demonstrated significant potential in generating high-quality synthetic data with realistic features. In this study, we leverage advanced diffusion models~\cite{rombach2022high} to produce 3,000 synthetic canine chest X-ray images, which are manually annotated with VHS scores. By augmenting the synthetic dataset with the original dataset, we enhance the model's robustness and diagnostic accuracy. This work underscores the transformative potential of AI, particularly diffusion-based models, in advancing veterinary diagnostic capabilities through the integration of synthetic data and pseudo-labeling.

%\noindent\textbf{Synthetic Data Augmentation and Benchmarking.} To enhance the original dataset, we generated a comprehensive collection of 3,000 high-quality synthetic images, each manually annotated with precise Vertebral Heart Score (VHS) key points. This augmented dataset significantly increases data diversity, enabling the model to better handle variability in anatomical structures and imaging conditions. By serving as a benchmark for evaluating model performance, the enriched dataset improves robustness and enhances diagnostic accuracy in detecting canine cardiomegaly.

%\noindent\textbf{Pseudo-labeling for Iterative Model Refinement.} We incorporate high-confidence pseudo-labels generated by the initial CDA model into the training set, enhancing model accuracy. To identify high-confidence labels, we employ Monte Carlo (MC) Dropout, which enables uncertainty estimation by performing multiple stochastic forward passes through the model during inference. Specifically, for each unlabeled sample, the model generates a series of predictions, from which we compute the mean prediction and standard deviation as measures of confidence. Only samples with uncertainty (standard deviation) below a predefined threshold (e.g., 0.005) are selected as high-confidence pseudo-labels. This process ensures the inclusion of reliable samples, minimizing the risk of introducing noise and demonstrating the effectiveness of semi-supervised learning in veterinary applications.

%-------------------------------------------------------------------------

\section{Related Work}

Thoracic radiography remains one of the most essential diagnostic tools for assessing cardiac diseases in veterinary medicine, particularly in detecting canine cardiomegaly.
\noindent\textbf{Vertebral Heart Score (VHS) Methods.} Traditionally, the vertebral heart score (VHS) is used to quantify heart enlargement, requiring manual measurement of the heart's short (S) and long (L) axes and determining the position of the fourth vertebra. The VHS is then calculated as the sum of these axes' lengths divided by the vertebral length. Although this method is widely utilized, studies have revealed significant limitations regarding its accuracy and consistency. For example, Rungpupradit et al.~\cite{Rungpupradit2020} identified variability when comparing conventional VHS methods to modified approaches in healthy Thai domestic shorthair cats, highlighting issues in traditional VHS reliability due to anatomical variation across species. Similarly, Tan et al.~\cite{Tan2020} observed that VHS measurements correlate with pulmonary patterns in dogs, though this correlation also exposed VHS's sensitivity to image quality and measurement inconsistency, underscoring the potential for improvement. Bappah et al.~\cite{Bappah2021} examined the relationship between VHS and cardiac sphericity, finding that manually labeled VHS is prone to human error, making it a time-intensive process lacking reproducibility. To address these challenges, Li et al.\cite{li2024regressive} introduced a novel regressive modeling framework leveraging deep learning techniques for automated VHS measurement, significantly improving efficiency and reducing the variability caused by manual input. 
% Their work demonstrated that automated approaches not only enhance reproducibility but also offer the potential for integrating additional predictive parameters, facilitating a more comprehensive assessment of cardiac health in veterinary practice.

\noindent\textbf{Deep Learning for VHS and Cardiomegaly Assessment.} To address these limitations, deep learning techniques, particularly convolutional neural networks (CNNs), have been introduced to assist and potentially replace traditional VHS methods in veterinary cardiology. Zhang et al.~\cite{zhang2021computerized} advanced the use of CNNs by calculating VHS based on key points detected by deep learning models, which were then cross-referenced with VHS standards across dog breeds, providing a more automated and precise approach to canine cardiomegaly evaluation. Additionally, Jeong and Sung~\cite{Jeong2022} proposed an adjusted heart volume index (aHVI) as a deep learning-based alternative to VHS, utilizing retrospective data to quantify heart size and further improve diagnostic accuracy. Similarly, Burti et al.~\cite{Burti2020} developed a computer-aided detection (CAD) tool based on CNNs to automate cardiomegaly detection, thus facilitating a more rapid and scalable diagnostic process. Dumortier et al.~\cite{Dumortier2022} expanded on these methods by adapting ResNet50V2 to classify feline thoracic radiographs, showing improved classification accuracy for pulmonary patterns in cats. These studies collectively highlight the adaptability of CNNs to a range of veterinary applications, though challenges remain in directly calculating VHS values via CNNs with sufficient interpretability and precision.

\noindent\textbf{Image Generation for Veterinary Diagnostics.} In parallel, image generation techniques, particularly diffusion models and generative adversarial networks (GANs), have emerged as powerful tools to augment limited datasets in medical imaging. Diffusion models~\cite{Ho2020, rombach2022high, yueh2025diffusion} have gained significant attention due to their ability to model complex data distributions, producing high-quality and diverse synthetic images. Unlike GANs, which were introduced by Goodfellow et al.\cite{Goodfellow2014} to generate realistic synthetic images, diffusion models iteratively refine data through a noise-reduction process, achieving superior results in many cases. For instance, diffusion models have been successfully applied to generate high-quality chest X-rays\cite{Pinaya2022}, as well as synthetic diagnostic images for veterinary medicine~\cite{yueh2025diffusion}, demonstrating their versatility in addressing domain-specific challenges. Yueh et al.~\cite{yueh2025diffusion} further highlight how diffusion models can enhance diagnostic accuracy by providing augmented datasets tailored to underrepresented conditions in veterinary diagnostics.

%Building on these advancements, our approach integrates synthetic image generation and pseudo-labeling to address the constraints of limited labeled veterinary data. By leveraging synthetic images generated from GANs and combining them with pseudo-labeled data, we aim to enhance model generalization and diagnostic accuracy for VHS prediction.

\noindent\textbf{Pseudo-labeling in Deep Learning.} Recently, pseudo-labeling has emerged as an effective semi-supervised learning technique to improve model performance, particularly when labeled data is limited~\cite{zhang2020adversarial1,zhang2022separable,zhang2022dairy,zhang2021weighted}. Lee et al.~\cite{Lee2013} introduced the concept of pseudo-labeling, demonstrating that incorporating high-confidence predictions from a model into the training process can improve generalization. Sohn et al.~\cite{Sohn2020} further refined this approach by integrating uncertainty estimation with pseudo-label selection, enabling models to effectively utilize unlabeled data while maintaining robustness. Applications of pseudo-labeling in medical imaging, such as X-ray classification and segmentation, have shown significant success. For example, Bai et al.~\cite{Bai2019} used pseudo-labels to enhance segmentation accuracy in cardiac MRI images, while Zhang et al.~\cite{Zhang2023} demonstrated its utility in improving lung disease detection on chest X-rays by iteratively refining model predictions.

Building on these advancements, our approach integrates synthetic image generation and pseudo-labeling to address the challenges of limited labeled veterinary data. We enhance model generalization and diagnostic accuracy for VHS prediction by combining synthetic images generated from diffusion models with high-confidence pseudo-labeled data. Pseudo-labeling further expands the effective training dataset in veterinary radiography tasks, where labeled data is often scarce, by leveraging high-confidence predictions from pre-trained models without requiring additional manual labeling effort.

%-------------------------------------------------------------------------

\section{Methodology}

\begin{figure*}[t]
    \centering
    \includegraphics[width=\textwidth]{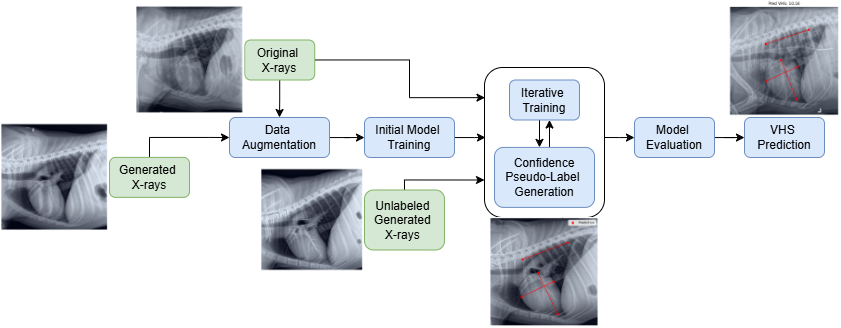}
    \caption{Overall framework of the Confident Pseudo-labeled Diffusion Augmentation (CDA) model. The framework integrates synthetic data augmentation, high-confidence pseudo-labeling, and iterative training for reliable VHS prediction.}
    \label{fig:frame}
\end{figure*}

\subsection{Confident Pseudo-labeled Diffusion Augmentation (CDA) model} 
We introduce the Confident Pseudo-labeled Diffusion Augmentation (CDA) model, a novel artificial intelligence (AI)-based approach for automating the prediction of Vertebral Heart Score (VHS) from canine chest X-rays. This methodology integrates three core components: synthetic data augmentation using diffusion models, pseudo-labeling with Monte Carlo Dropout for leveraging unlabeled data, and a tailored neural network architecture optimized for this task. Together, these strategies address critical challenges such as the scarcity of annotated datasets and variability in X-ray imaging quality, resulting in a robust and reliable predictive model. \cref{fig:frame} provides an overview of the CDA framework, showcasing the integration of data augmentation, high-confidence pseudo-labeling, and iterative training. This systematic approach ensures high generalizability and accuracy across diverse imaging conditions.

% \subsection{Model Architecture}
%The base model architecture utilized was EfficientNet-B7, chosen for its balance of computational efficiency and performance, leveraging a compound scaling method to adjust its depth, width, and resolution. To adapt EfficientNetB7 for our task, we replaced the default classifier head with a custom augmentation head designed to output a 12-class prediction, reflecting the specific VHS classification used in this study. 

%The AugmentHead class defines a linear layer with 12 output units, representing the categories for different VHS scores. This custom layer takes the feature representation from EfficientNetB7's classifier and projects it onto the desired classification space.

%The AugmentHead is then integrated into EfficientNetB7, replacing its default classification layer. This modified architecture provides greater flexibility in classifying VHS scores with multiple class distinctions, aiding in more accurate cardiomegaly detection.

%%%%%%%%%%%%%%%%%%%%%%%%%%%%%%%%%%%%

%\subsection{Initial Model Training}

To address the scarcity of labeled data, we generate synthetic X-ray images using advanced diffusion models~\cite{rombach2022high}. These models are trained on an original  dataset of canine chest X-rays to produce anatomically plausible synthetic images to reflect the variations observed in real-world clinical settings. Then, we manually annotate them with precise Vertebral Heart Score (VHS) values, ensuring high-quality labels. This augmentation process significantly enriches the diversity of the training dataset, particularly for rare or extreme cases of canine cardiomegaly, thereby improving the model's robustness and generalization capabilities.

In addition, We use EfficientNet-B7~\cite{tan2019efficientnet} as the base model architecture, chosen for its balance of computational efficiency and performance through compound scaling of depth, width, and resolution. To adapt it for this study, we replace the default classifier head with a custom augmentation head, which consists of a linear layer with 12 output units representing the VHS score categories. This AugmentHead projects the feature representation from EfficientNet-B7's classifier onto the desired classification space, allowing for more accurate cardiomegaly detection with multiple class distinctions. We use this model to predict the VHS score and to classify heart size into three clinically significant categories based on the VHS range. Specifically, heart size is categorized as follows:
\begin{eqnarray}\label{eq:y}
y_t = \begin{cases} 
      0 & \text{if } \text{VHS} < 8.2, \\
      1 & \text{if } 8.2 \leq \text{VHS} \leq 10, \\
      2 & \text{if } \text{VHS} > 10. 
   \end{cases}
\end{eqnarray}

\paragraph{Loss Function.} To achieve these objectives, we employ a composite loss function that integrates multiple components to enhance both regression accuracy and stability during training. The overall loss function is defined as:
\begin{align}
    \mathcal{L} = \frac{1}{n} \sum_{i=1}^n &\left( 10 \cdot \mathcal{L}_{\text{L1}}(f(x_i), y_i^{\text{points}}) \right. \nonumber \\
    & + 0.1 \cdot \mathcal{L}_{\text{L1}}(\text{calc\_vhs}(f(x_i)), y_i^{\text{VHS}}) \nonumber \\
    & \left. + \mathbb{1}_{\{ \text{epoch} > 10 \}} \cdot \mathcal{L}_{\text{L1}}(f(x_i), y_i^{\text{soft}}) \right), \label{eq:loss}
\end{align}
where \(\mathcal{L}_{\text{L1}}\) represents the mean absolute error (L1 loss),  \(f(x_i)\) denotes the model output for input \(x_i\), \(y_i^{\text{points}}\) is the target key points for each image, \(y_i^{\text{VHS}}\) is the VHS ground truth calculated from \(y_i^{\text{points}}\), \(y_i^{\text{soft}}\) is the averaged soft label from historical predictions, applied after the 10th epoch and \(\mathbb{1}_{\{ \text{epoch} > 10 \}}\) is an indicator function that activates the soft label loss term only if the epoch is greater than 10. This multi-term formulation balances multiple objectives to ensure stability and accuracy during training. Inspired by Zheng et al.~\cite{zheng2020error}, potential label noise is mitigated by selectively activating historical soft labels.

\paragraph{VHS Calculation.} The VHS, a key metric for assessing cardiomegaly, is calculated based on specific anatomical landmarks within the X-ray image. The computation is as follows:
\begin{itemize}
    \item Let \(A\), \(B\), \(C\), \(D\), \(E\), and \(F\) represent specific anatomical points within the X-ray, predicted by the model output \(f(x)\). 

    \item The distance between each pair of points is calculated as the Euclidean norm (L2 distance):
   
    \item The VHS is then computed using the formula:
    \begin{equation}
    \text{VHS} = 6 \cdot \frac{AB + CD}{EF},
    \end{equation}
    where \(AB\) and \(CD\) represent the long and short axes of the heart, and \(EF\) represents the length of the vertebral segment. The factor of 6 is applied to scale the ratio in line with veterinary standards for VHS measurement.
\end{itemize}

\begin{figure}[h]
    \centering
    \includegraphics[width=8cm]{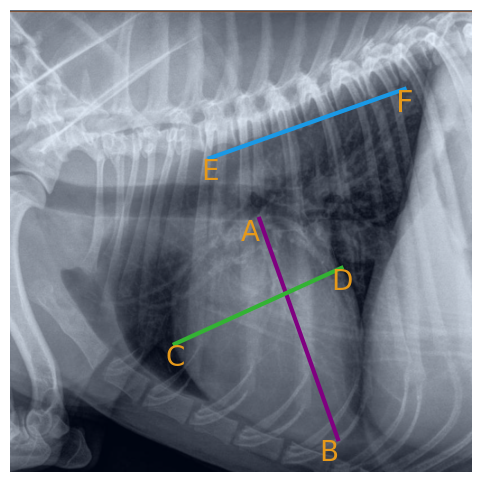}
    \caption{Illustration of anatomical landmarks for VHS calculation. \(AB\) (long axis), \(CD\) (short axis), and \(EF\) (vertebral segment) are shown, which are used to compute the Vertebral Heart Score (VHS).}
    \label{fig:vhs}
\end{figure}

The anatomical landmarks \(AB\), \(CD\), and \(EF\), used for VHS computation, are visually illustrated in~\Cref{fig:vhs}. Here, \(AB\) represents the heart's long axis, \(CD\) is the short axis, and \(EF\) corresponds to the vertebral segment, which provides the normalization factor.

%To balance regression and classification objectives, the primary loss terms were weighted at 10 (key point prediction) and 0.1 (VHS regression). The delayed introduction of the soft label loss term (\(y_i^{\text{soft}}\)) after 10 epochs allowed the model to first stabilize its anatomical predictions before incorporating historical averages for fine-tuning.

The overall training algorithm is provided in Alg.~\ref{alg:train}.
\begin{algorithm}
\caption{Training initial CDA Model for Canine Cardiomegaly Assessment}\label{alg:train}
\begin{algorithmic}[1]
\REQUIRE Canine X-ray images \( X = \{x_i\}_{i=1}^n \) and corresponding VHS-labeled key points \( P = \{y_i\}_{i=1}^n \), where \( n \) is the total number of images.
\ENSURE Predicted VHS scores and classifications for heart size.

\FOR{epoch = 1 to num\_epochs}
    \FOR{each mini-batch \( B(X) \) and \( B(P) \) sampled from \( X \) and \( P \)}
        \STATE Compute true VHS scores and classification labels using Eq.~\eqref{eq:y}.
        \STATE Apply model \( f \) to predict VHS scores and heart size classes.
        \STATE Calculate total loss \( \mathcal{L} \) using Eq.~\eqref{eq:loss}.
        \STATE Update model parameters via backpropagation and optimizer step.
    \ENDFOR
\ENDFOR

\RETURN Final trained model for VHS prediction and cardiomegaly classification.
\end{algorithmic}
\end{algorithm}

% \paragraph{Results}
% By the end of training, the model achieved a test accuracy of 91\%, demonstrating reliable VHS prediction and classification capabilities. This phase established a strong foundation for subsequent refinements through pseudo-labeling.

%%%%%%%%%%%%%%%%%%%%%%%%%%%%%%%%%%%%

\subsection{Pseudo-Labeling with Monte Carlo Dropout}

We employ pseudo-labeling to leverage the abundant unlabeled canine chest X-rays, enabling further improvements in model performance. This approach integrates predictions from a pre-trained model with uncertainty estimation to create high-confidence pseudo-labels, which are subsequently used to expand the labeled training dataset. % \paragraph{Monte Carlo Dropout for Uncertainty Estimation}
To estimate uncertainty in the model's predictions, Monte Carlo (MC) Dropout~\cite{gal2016dropout} is applied during inference. By enabling dropout layers during prediction, the model generates multiple stochastic outputs for the same input. This allows the computation of both the mean prediction and the associated uncertainty as the standard deviation of the predictions. For a given image \(x\), the stochastic predictions are denoted as:
\[
\{f_k(x)\}_{k=1}^{K},
\]
where \(f_k(x)\) represents the prediction from the \(k\)-th stochastic forward pass, and \(K\) is the total number of passes. The mean prediction and uncertainty are calculated as:
\begin{align}
\mu(x) &= \frac{1}{K} \sum_{k=1}^{K} f_k(x), \label{eq:mean} \\
\sigma(x) &= \sqrt{\frac{1}{K} \sum_{k=1}^{K} \left(f_k(x) - \mu(x)\right)^2}, \label{eq:stddev}
\end{align}
where \(\mu(x)\) represents the pseudo-label, and \(\sigma(x)\) represents the model's uncertainty.

% \paragraph{High-Confidence Pseudo-Labels}

To ensure the reliability of the pseudo-labels, a confidence threshold is applied to filter out uncertain predictions. Specifically, pseudo-labels with uncertainties below a predefined threshold \(\tau\) are considered high confidence:
\[
\mathcal{C} = \{x \in \mathcal{U} \mid \max(\sigma(x)) < \tau\},
\]
where \(\mathcal{U}\) is the unlabeled dataset and \(\mathcal{C}\) denotes the subset of high-confidence samples.

The selected high-confidence pseudo-labels are then combined with the labeled dataset to create an expanded training dataset for the next training iteration.

\paragraph{Loss Function for Pseudo-Labeling}

The loss function for pseudo-labeling incorporates both the labeled and high-confidence pseudo-labeled datasets. Let \(\mathcal{L}_{\text{L1}}\) denote the L1 loss, and let \(\mathcal{D}_{\text{Labeled}}\) and \(\mathcal{D}_{\text{Pseudo}}\) represent the labeled and pseudo-labeled datasets, respectively. The total loss is expressed as:
\begin{align}
\mathcal{L}_{\text{Total}} &= \frac{1}{|\mathcal{D}_{\text{Labeled}}|} \sum_{(x_i, y_i) \in \mathcal{D}_{\text{Labeled}}} \mathcal{L}_{\text{L1}}(f(x_i), y_i) \nonumber \\
&+ \lambda \cdot \frac{1}{|\mathcal{D}_{\text{Pseudo}}|} \sum_{(x_i, \mu_i) \in \mathcal{D}_{\text{Pseudo}}} \mathcal{L}_{\text{L1}}(f(x_i), \mu_i),
\end{align}
where \(\lambda\) is a weighting factor to balance the contributions of the labeled and pseudo-labeled datasets.

% \paragraph{Training Algorithm}
We integrate the pseudo-labeling process into the training pipeline, iteratively expanding the dataset and improving the model's performance. By iteratively applying pseudo-labeling with MC Dropout, the model's accuracy on the test dataset improved from 91\% (after initial training) to 92.75\%, demonstrating the efficacy of this approach in utilizing unlabeled data to enhance predictive performance. The training algorithm is detailed in Alg.~\ref{alg:pseudo}.

\begin{algorithm}
\caption{Pseudo-Labeling with Monte Carlo Dropout}\label{alg:pseudo}
\begin{algorithmic}[1]
\REQUIRE Pre-trained model \(f\), labeled dataset \(\mathcal{D}_{\text{Labeled}}\), unlabeled dataset \(\mathcal{D}_{\text{Unlabeled}}\), confidence threshold \(\tau\), number of MC Dropout passes \(K\).
\ENSURE Trained model with improved performance.

\FOR{epoch = 1 to num\_epochs}
    \STATE Train \(f\) on \(\mathcal{D}_{\text{Labeled}}\) using the current loss function.
    \STATE Enable MC Dropout during inference for \(\mathcal{D}_{\text{Unlabeled}}\).
    \FOR{each image \(x \in \mathcal{D}_{\text{Unlabeled}}\)}
        \STATE Perform \(K\) stochastic forward passes to obtain \(\{f_k(x)\}_{k=1}^K\).
        \STATE Compute mean prediction \(\mu(x)\) and uncertainty \(\sigma(x)\) using Eqs.~\eqref{eq:mean} and ~\eqref{eq:stddev}.
    \ENDFOR
    \STATE Select high-confidence samples \(\mathcal{C}\) based on \(\max(\sigma(x)) < \tau\).
    \STATE Combine \(\mathcal{C}\) with \(\mathcal{D}_{\text{Labeled}}\) to create an expanded training dataset.
\ENDFOR

\RETURN Trained model \(f\).
\end{algorithmic}
\end{algorithm}

% \paragraph{Results}

%%%%%%%%%%%%%%%%%%%%%%%%%%%%%%%%

\subsection{Datasets}

\subsubsection{Data Collection}

The dataset utilized in this study builds upon the original DogHeart dataset, which consists of X-ray images of canine thoracic radiographs from Shanghai Aichong Pet Hospital, as documented in the original paper~\cite{li2024regressive}. The initial dataset includes 6,389 images, with 1,400 used for training, 200 for validation, and 400 for testing. We pre-process these images by cropping to focus solely on the thoracic region, ensuring privacy by removing any identifying information. We expanded this DogHeart dataset by generating an additional 3,000 synthetic images to increase diversity and improve model generalization. Each synthetic image is manually labeled to match the standards of the original datasets, using VHS scores to categorize the images into three groups: small hearts (VHS $<$ 8.2), normal hearts (VHS between 8.2 and 10), and large hearts (VHS $>$ 10). After augmentation, the dataset consists of 5,000 images, with 4,400 (88\%) for training, 200 (4\%) for validation, and  400 (8\%) for testing. This expanded dataset allows for more robust training,by providing a broader range of samples, helping reduce overfitting and improve generalization.

\subsubsection{Data Labeling}

Accurate labeling is critical to the performance of deep learning models, especially in medical imaging tasks such as VHS prediction. To ensure high-quality annotations, we used the specialized dog heart analysis software developed in~\cite{li2024regressive}. 
%
% This software facilitates the precise annotation of six key points in each X-ray image, which are essential for calculating the VHS score. The software includes the following functionalities:
% \begin{itemize}
%     \item Manual Point Labeling: Users can load X-ray images and place six key points, which correspond to anatomical landmarks used to calculate the VHS score. The software overlays these points onto the X-ray images, allowing visual verification and alignment.
%     \item Data Saving and Verification: Annotated points can be saved for each image, and the software provides a ‘Saved’ folder to organize and store all verified points. This feature supports efficient data management, especially for large datasets like ours.
%     \item Performance Comparison: The software also allows for the comparison of ground truth annotations with model-predicted points, enabling quick evaluation of the model's performance in accurately identifying anatomical landmarks.
% \end{itemize}
The software follows a standardized protocol for VHS measurement. For each image, the four key points defining the long and short axes of the heart are identified first. The long axis is defined from the carina to the apex of the heart, while the short axis spans the widest part of the heart and is perpendicular to the long axis. Two additional points mark the fourth thoracic vertebra (T4) and the ninth vertebra, allowing for a consistent VHS calculation. 

We applied the same rigorous labeling process to the synthetic images, ensuring that the newly generated data adhered to the standards set by the original dataset. This manual annotation process is both time-intensive and costly; however, it is essential for producing a reliable dataset that accurately represents the VHS values across diverse samples. Each image, whether real or synthetic, was labeled by experts following the same protocol, guaranteeing consistent and reliable ground truth annotations.

% \subsubsection{Dataset Statistics}

% The expanded dataset includes three categories of VHS scores, reflecting the size of the canine heart: Small Hearts: VHS scores below 8.2; Normal Hearts: VHS scores between 8.2 and 10; and Large Hearts: VHS scores above 10. The synthetic data augmentation provided a more balanced distribution across categories, especially beneficial for the small heart category, which had fewer samples in the original dataset.

\subsection{Implementation details}
The model training process was conducted over 1000 epochs using the AdamW optimizer, which combines adaptive learning with weight decay regularization to mitigate overfitting. A cosine annealing learning rate scheduler was employed to dynamically adjust the learning rate, ensuring efficient convergence. A batch size of 16 was used, and gradient accumulation was implemented to simulate larger batch sizes without exceeding hardware memory limitations. The training loop incorporated regular validation to monitor model performance, using metrics such as validation loss and accuracy based on the classification thresholds defined in Eq.~\eqref{eq:y}.

%%%%%%%%%%%%%%%%%%%%%%%%%%%%%%%%

\begin{figure*}[t]
    \centering
    \includegraphics[width=\textwidth]{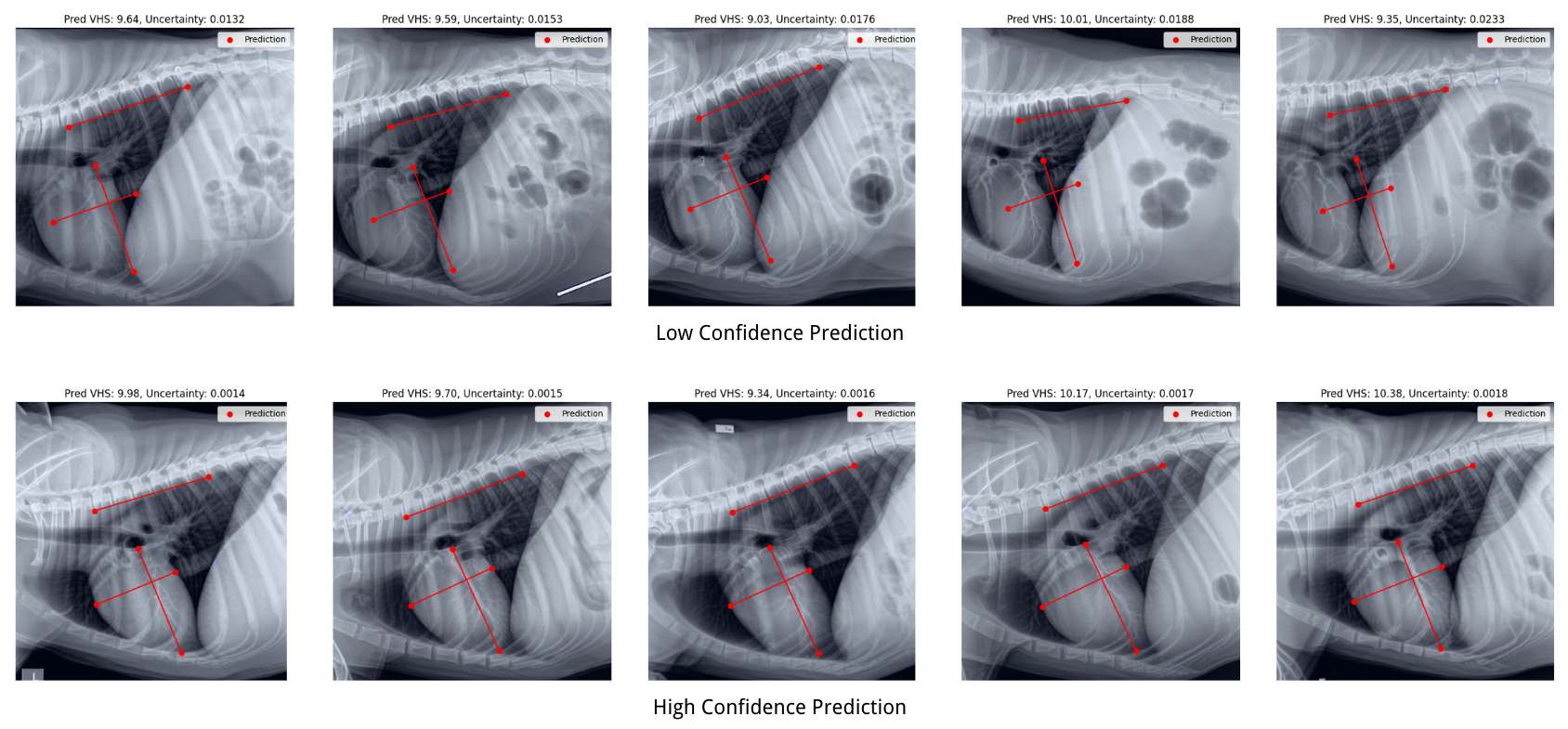}
    \caption{Comparisions of high confidence prediction and low confidence presiction.}
    \label{fig:conf}
\end{figure*}

\begin{figure*}[t]
    \centering
    \includegraphics[width=\textwidth]{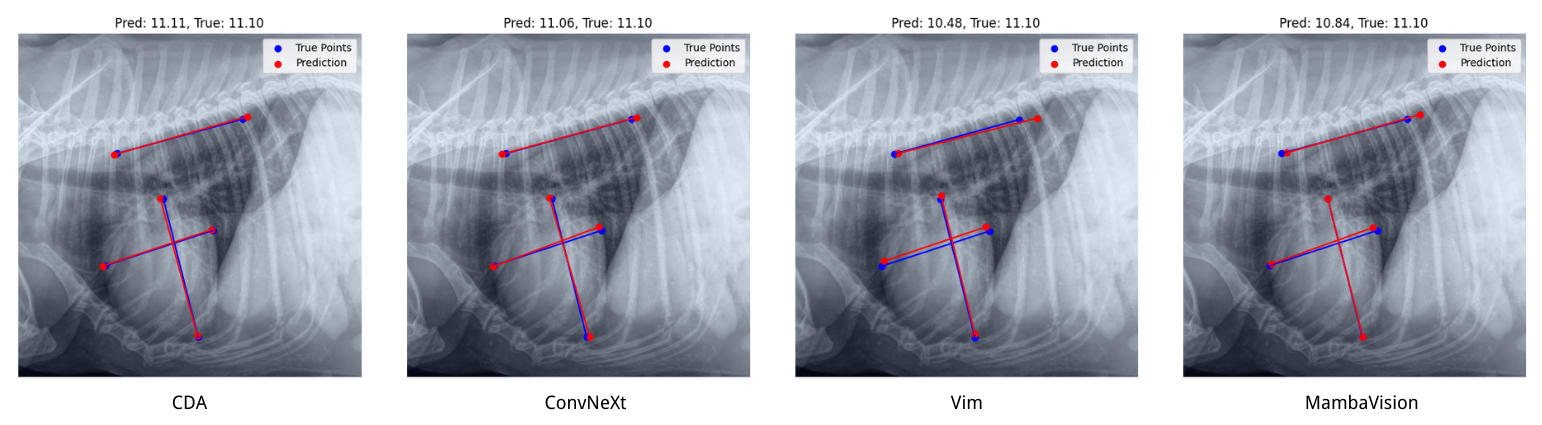}
    \caption{Prediction comparisons of different models.}
    \label{fig:results}
\end{figure*}

\section{Results}

This section presents the experimental results of our approach, providing a detailed evaluation of its effectiveness in predicting Vertebral Heart Score (VHS) and classifying canine cardiomegaly. The primary evaluation metric for this study is the test accuracy of VHS-based classification into three clinically significant categories: normal heart size (\( \text{VHS} < 8.2 \)), borderline cardiomegaly (\( 8.2 \leq \text{VHS} \leq 10 \)), and severe cardiomegaly (\( \text{VHS} > 10 \)). These categories provide clear thresholds for assessing the presence and severity of cardiomegaly, which are crucial in veterinary diagnostics.

~\Cref{tab:accuracy_comparison_pseudo} compares the performance of various state-of-the-art models and highlights the advantages of our proposed Confident Pseudo-labeled Diffusion Augmentation (CDA) model. Notably, CDA achieves a test accuracy of 92.8\%, outperforming other baselines, including ConvNeXt (89.8\%), MambaVision (86.8\%), and ViM (71.5\%). Without the inclusion of pseudo-labeled data, the CDA model achieves 91.0\%, demonstrating that pseudo-labeling contributes significantly to performance gains by leveraging high-confidence labeled samples, as shown in~\Cref{fig:conf}

\begin{table}[t]
\small
\begin{center}
% \vspace{-0.3cm}
%\captionsetup{font=small}
%\caption{Results comparisons of different methods (Acc: accuracy, CPL: Confident Pseudo labels). }
\setlength{\tabcolsep}{+0.3cm}{
\begin{tabular}{lcclllllllllll}
\hline 
\textbf{Model}  & \textbf{Valid Acc} & \textbf{Test Acc} \\
\hline
GoogleNet~\cite{szegedy2015going}    & 77.5  & 74.8\\
VGG16~\cite{simonyan2014very}       & 78.5  & 75.0\\
ResNet50~\cite{he2016deep}     &80.0  &78.2 \\
DenseNet201~\cite{huang2017densely}     & 77.0  & 80.8\\
Inceptionv3~\cite{szegedy2016rethinking}     & 79.0  & 80.0\\
Xception~\cite{chollet2017xception}      & 78.5 & 75.2 \\
InceptionResnetV2~\cite{szegedy2017inception}      &  77.5 & 78.8\\
NasnetLarge~\cite{zoph2018learning}      & 80.0  & 82.5\\
EfficientNetB7~\cite{tan2019efficientnet}      & 82.0  &84.5 \\
Vision transformer~\cite{dosovitskiy2020image}       & 80.0  & 77.5\\
CONVT~\cite{wu2021cvt}      & 82.0  & 85.3\\
Beit\_large~\cite{bao2021beit}     & 71.0  &74.3 \\
RVT~\cite{li2024regressive}     & 84.9  & 87.3 \\
ConvNeXt~\cite{liu2022convnet}    &89.5 &89.8\\
Vim~\cite{zhu2024vision} &73.5 &71.5\\
MambaVision~\cite{hatamizadeh2024mambavision} &87.5 &86.8\\
\hline  
\hline
\textbf{CDA w/o CPL }     & 88.5  & 91.0 \\
\textbf{CDA}     & \textbf{89.5 } & \textbf{92.8} \\
\hline
\end{tabular}}
\captionsetup{font=small}
\caption{Results comparisons of different methods (Acc: accuracy, CPL: Confident Pseudo labels). }
\label{tab:accuracy_comparison_pseudo}
\end{center}
\end{table}

To further illustrate the performance of CDA compared to other methods, ~\Cref{fig:results} visually compares the predicted key points and VHS values across multiple models. The CDA model demonstrates superior alignment of the predicted key points (red) with the true anatomical key points (blue), achieving a predicted VHS value closest to the ground truth. In contrast, other models such as ConvNeXt, ViM, and MambaVision show varying degrees of deviation, particularly in key point alignment and VHS estimation accuracy.

The results highlight the clinical relevance of our approach. By achieving a high test accuracy of 92.8\%, CDA demonstrates its robustness and reliability in VHS prediction and cardiomegaly classification. This level of accuracy is particularly critical for reducing misclassifications in borderline cases, which are often challenging to diagnose. The integration of confident pseudo-labeled data played a key role in enhancing model generalization, as it effectively leveraged high-confidence samples from unlabeled data without introducing noise. 

In summary, our experimental results validate the effectiveness of the proposed CDA model. The incorporation of diffusion-based synthetic data augmentation and pseudo-labeling strategies significantly improved both the accuracy and generalization of the model, setting a new benchmark for VHS prediction in canine cardiomegaly diagnostics.

\subsection{Ablation study}

\noindent\textbf{Impact of Pseudo-Labeling. } To evaluate the contribution of pseudo-labeling, we analyzed the quality and quantity of pseudo-labeled samples included in the training dataset. Using Monte Carlo (MC) Dropout~\cite{gal2016dropout}, uncertainty was estimated for each unlabeled sample, and only high-confidence predictions (uncertainty below the threshold of 0.005) were incorporated.  The pseudo-labeling process dynamically expanded the training dataset with high-confidence samples, leading to improved model generalization. By filtering out low-confidence predictions, we ensured that the pseudo-labeled samples contributed positively to the learning process without introducing significant noise.

~\Cref{fig:conf} compares predictions with high and low uncertainty. The top row of the figure shows examples of low-confidence predictions, where the uncertainty values exceed the threshold, often leading to noticeable deviations in the predicted anatomical key points. In contrast, the bottom row presents high-confidence predictions with uncertainties below the threshold. These predictions demonstrate greater accuracy, as indicated by the well-aligned key points and consistent Vertebral Heart Score (VHS) measurements.  

By filtering out low-confidence predictions, the pseudo-labeling process ensures that only reliable samples are included in the training set, improving the model's robustness and generalization.

%%%%%%%%%%%%%%%%%%%%%%%%%%%%%%%%%%%%%%%%%%%%%%%%%%%%%%%%%%%%%%%%%%%%%%%%

\section{Discussion}

We explore the use of advanced deep learning techniques, including pseudo-labeling and synthetic data augmentation, to improve the prediction of Vertebral Heart Score (VHS) and the classification of canine cardiomegaly from chest X-rays. The experimental results demonstrate the effectiveness of these methods, achieving a test accuracy of 92.75\% and highlighting their potential in veterinary diagnostics. This section discusses the broader implications of these findings, the advantages and limitations of the proposed approach, and opportunities for future research.

%%%%%%%%%%%%%%%%%%%%%%%%%%%

\textbf{Advantages of Data Augmentation.}
The integration of synthetic images in this study offers several key benefits. First, it mitigates the challenges posed by limited training data in deep learning by expanding the dataset with a diverse set of high-quality images, thereby enhancing the model's ability to generalize to unseen cases. Second, synthetic data introduces controlled variations that encourage the model to learn more nuanced and complex features, improving its accuracy in predicting Vertebral Heart Scores (VHS) across both real and synthetic images. Finally, data augmentation effectively addresses class imbalance, a common issue in medical datasets, by generating rare or extreme cases, such as severe heart enlargement or reduction, ensuring these underrepresented categories are adequately learned and represented.

%%%%%%%%%%%%%%%%%%%%%%%%%%%%%%

\textbf{Advantages of Pseudo-Labeling.} Pseudo-labeling demonstrated substantial improvements in our model's performance, as evidenced by the results obtained in this study. By leveraging unlabeled canine chest X-ray images, the pseudo-labeling process expanded the training dataset with high-confidence predictions, effectively overcoming the limitations of limited labeled data. As shown in ~\Cref{fig:conf}, high-confidence samples, selected using Monte Carlo (MC) Dropout~\cite{gal2016dropout} with an uncertainty threshold of 0.005, exhibited accurate anatomical key point predictions and consistent Vertebral Heart Score (VHS) measurements. This ensured that the additional training examples introduced minimal noise while enhancing the model's learning capacity.

The incorporation of high-confidence pseudo-labeled samples significantly improved the model's generalization ability. The enhanced variability in anatomical structures and imaging conditions enabled the model to adapt to real-world scenarios, where such variations are common. Compared to low-confidence predictions, which often introduced misalignments and greater uncertainty, filtering these samples ensured robust training. This process directly contributed to the observed 1.75\% improvement in test accuracy, demonstrating the practical impact of pseudo-labeling on the CDA model's ability to reliably predict VHS and classify cardiomegaly. These results validate that pseudo-labeling not only mitigates data scarcity challenges but also provides a mechanism to incorporate diverse, high-quality examples, ultimately leading to superior predictive performance.

\textbf{Challenges.} While promising, the proposed approach has several limitations. The quality of synthetic data generated by GANs is critical; poorly generated images could introduce biases and affect model performance. Additionally, the fixed confidence threshold for pseudo-label selection may exclude borderline samples, which might otherwise enrich the training set. Adaptive thresholding could help address this issue. The training pipeline's computational requirements are another challenge. MC Dropout and iterative pseudo-labeling demand significant resources, potentially limiting applicability in resource-constrained settings. Simplifying the process could increase its accessibility. Finally, the lack of external validation limits the generalizability of the results. Evaluating the model on independent datasets or in real-world clinical settings is essential to confirm its robustness and broader applicability.

\textbf{Opportunities for Future Research.} Future work could focus on adaptive pseudo-labeling techniques to better balance confidence thresholds and sample inclusion. Advanced generative models, such as Latent Diffusion Models (LDMs)~\cite{rombach2022high} and their recent extensions for improved efficiency and diversity~\cite{selim2023diffusionct, moris2024adapted, prusty2024enhancing}, could further improve the quality and diversity of synthetic data, enhancing model generalization. Exploring cross-modality learning by integrating X-ray images with clinical notes or other imaging data offers another promising direction. This could provide richer contextual information, improving predictive accuracy. Additionally, developing explainable AI methods to visualize the model’s decision-making process would increase trust among clinicians, facilitating real-world adoption. External validation on diverse datasets is also critical to establish the robustness of the proposed approach and extend its applicability to broader clinical and veterinary use cases.

%%%%%%%%%%%%%%%%%%%%%%%%%%%%%%%%%%%%%%%%%%%%%%%%%%%%%%%%%%%%%%%%%%%%%%%%

\section{Conclusion}

We present a novel approach for predicting Vertebral Heart Score (VHS) and classifying canine cardiomegaly using deep learning techniques. By integrating pseudo-labeling with Monte Carlo Dropout and synthetic data augmentation, the proposed method overcomes the challenge of limited labeled datasets in veterinary imaging. The experimental results highlight the effectiveness of this approach, achieving a test accuracy of 92.75\% and outperforming state-of-the-art methods. These findings highlight the potential of this approach to advance veterinary diagnostics, enabling accurate VHS prediction and cardiomegaly classification for early detection and timely intervention, ultimately improving clinical outcomes for canines. 
% The methodology's adaptability and scalability also make it a promising tool for similar applications in other medical imaging domains.

% The study's main contributions can be summarized as follows:
% \paragraph{Innovative Semi-Supervised Learning Framework} The integration of pseudo-labeling enables the effective utilization of unlabeled data, dynamically expanding the training dataset with high-confidence pseudo-labeled samples. This approach enhances model generalization and robustness in predicting VHS.

% \paragraph{Synthetic Data Augmentation} The use of GAN-generated synthetic X-rays introduces variability into the training process, mitigating overfitting and ensuring the model's ability to handle diverse clinical scenarios.

% \paragraph{Comprehensive Evaluation} Through rigorous testing and analysis, the study demonstrates the impact of pseudo-labeling and synthetic augmentation on model performance, providing valuable insights into their benefits and limitations.

%%%%%%%%% REFERENCES
{\small
\bibliographystyle{ieee_fullname}
\bibliography{egbib}
}

\end{document}